\documentclass[a4paper,twoside]{article}

\usepackage{epsfig}
\usepackage{subfigure}
\usepackage{calc}
\usepackage{amssymb}
\usepackage{amstext}
\usepackage{amsmath}
\usepackage{amsthm}
\usepackage{multicol}
\usepackage{pslatex}
\usepackage{apalike}
\usepackage{SciTePress}
\usepackage[small]{caption}

\begin{document}

\title{Exploring the Potential of Combining Time of Flight and Thermal Infrared Cameras for Person Detection} 

\author{\authorname{Wim Abbeloos and Toon Goedem\'{e}}
\affiliation{KU Leuven, Department of Electrical Engineering, EAVISE\\Leuven, Belgium}
\email{wim.abbeloos@kuleuven.be, toon.goedeme@kuleuven.be}
}


\abstract{Combining new, low-cost thermal infrared and time-of-flight range sensors provides new opportunities.  In this position paper we explore the possibilities of combining these sensors and using their fused data for person detection.  The proposed calibration approach for this sensor combination differs from the traditional stereo camera calibration in two fundamental ways.  A first distinction is that the spectral sensitivity of the two sensors differs significantly.  In fact, there is no sensitivity range overlap at all.  A second distinction is that their resolution is typically very low, which requires special attention.  We assume a situation in which the sensors' relative position is known, but their orientation is unknown.  In addition, some of the typical measurement errors are discussed, and methods to compensate for them are proposed.  We discuss how the fused data could allow increased accuracy and robustness without the need for complex algorithms requiring large amounts of computational power and training data.}

\onecolumn \maketitle \normalsize \vfill

\section{\uppercase{Introduction}}
\label{sec:introduction}

\noindent Cameras have been used to record and monitor people's activities in a great variety of situations.  They provide an easy, affordable and intuitive way to observe our surroundings.  The automatic detection of people has important applications in the areas of machine safety, human-computer interaction, security, traffic analysis, driver assistance, health-care, etc.

Detecting people in images, however, turns out to be a surprisingly difficult task.  The major problem when detecting people is the immense variance in their appearance.  Let's just consider a few causes:

\begin{itemize}
    \item[\textperiodcentered] Intra-class variety: all people are unique.  We all have different body proportions, wear different clothes and move in a different way.
    \item[\textperiodcentered] The illumination conditions are often uncontrolled.  They may be completely unknown, or vary in time.
    \item[\textperiodcentered] A person's appearance strongly depends on the point of view.
    \item[\textperiodcentered] When using a regular camera, dimensional information is lost by projection.  The size of a person in the image depends on its distance to the camera.
    \item[\textperiodcentered] Articulateness: the human body is highly flexible.  Especially the limbs can take a large variety of poses.
    \item[\textperiodcentered] Often a person is only partially visible.  For example, when entering or leaving the cameras' field of view, or when occluded by other objects.
\end{itemize}

\noindent Despite these issues, some very powerful computer vision algorithms for the detection of people from normal camera images exist.  A lot of progress has recently been made in the detection of pedestrians \cite{Dollar2011}\cite{Enzweiler2008}.  Many of these algorithms use a Histogram of Oriented Gradients-based detector, combined with a part based model.  While the performance of these algorithms continues to improve, they require a lot of computational power and a very large annotated dataset for the training stage. They also rely on some situation specific assumptions (e.g. only people with an approximately vertical pose are detected).

An alternative approach, which avoids many of these issues, is not to detect people specifically, but to detect any moving object.  Especially in applications with a static camera, this can be done very easily and efficiently by applying background subtraction algorithms.  Methods such as approximate median filtering \cite{McFarlane1995} and shadow detection \cite{Rosin1995} can be used to increase the robustness to varying light conditions.

This approach has been used as a preprocessing step for pedestrian detection algorithms in order to segment the image, reducing the search space and thus the required processing time \cite{Liang2012}.

We explore a similar approach, but instead of measuring the amount of reflected light from an object, as is observed with a normal camera, we propose a sensor measuring thermal radiation and range.  Measuring these physical properties directly provides far more informative data, being the temperature of an object and its geometrical measures, respectively \cite{Gandhi2007}.  We anticipate that fused Time-of-Flight (TOF) range measurements and thermal infrared (IR) data will allow significant improvements in three key areas:
\begin{enumerate}
	\item{Accurate and fast segmentation of moving objects.}
	\item{Reduced complexity of people detection algorithms.}
	\item{Reduction of the required amount of training data.}
\end{enumerate}

\begin{figure}
  \vspace{-0.2cm}
  \centering
   {\epsfig{file = 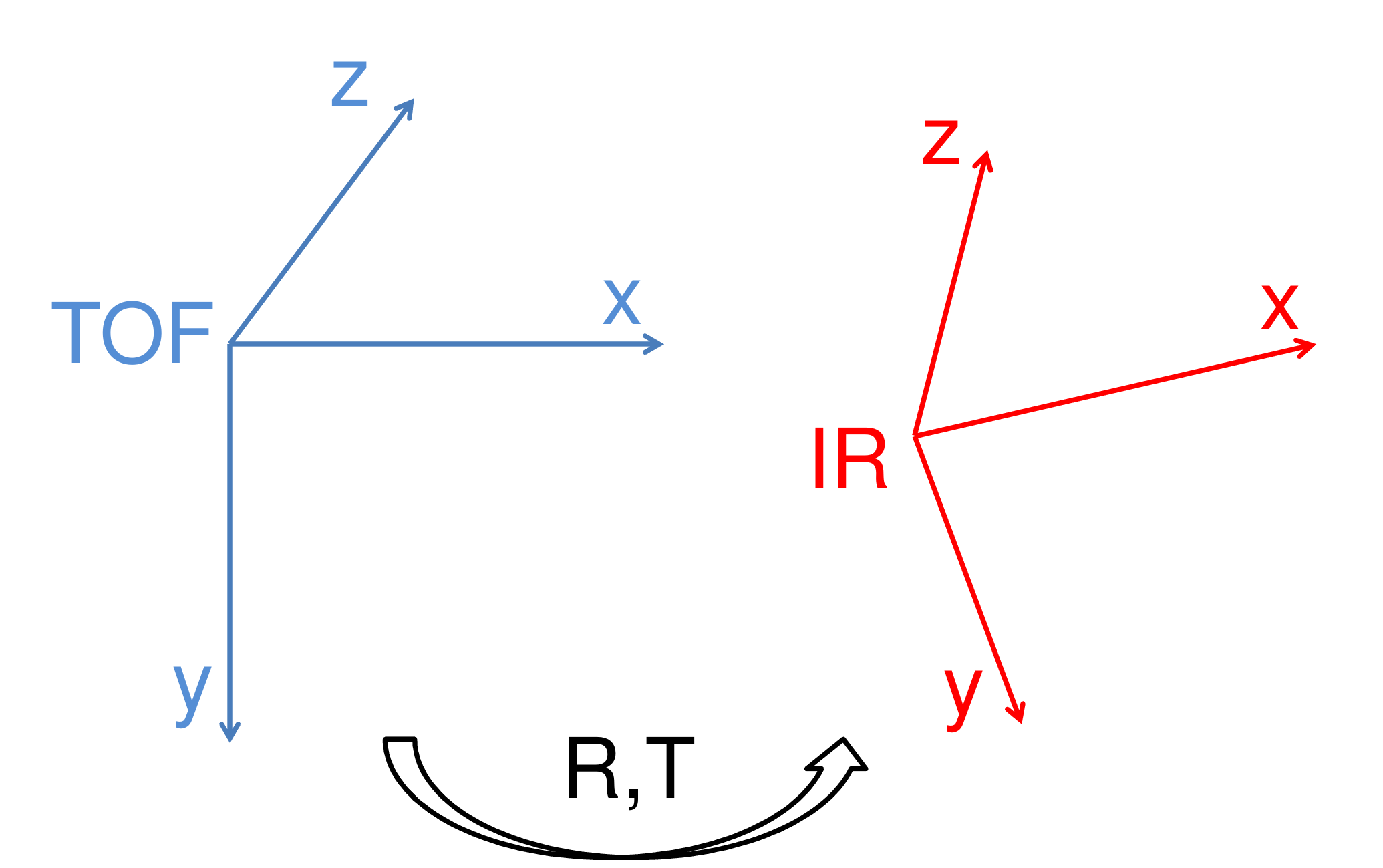, width = 4.5cm}}
  \caption{Relative pose of the TOF and IR camera.  The translation (T) is known, but the rotation (R) is unknown and must be determined during the calibration step.}
  \label{fig:coord}
  \vspace{-0.1cm}
\end{figure}

\noindent The following sections provide more details on these sensors.  A prototype combining a TOF and IR camera is currently being developed.  The relative translation of these sensors is known sufficiently accurately, but their relative rotation is not (figure \ref{fig:coord}).  To fuse their data and obtain a 3D thermogram we propose calibration routine, described in sections \ref{sec:fusion} and \ref{sec:cal}.  Preliminary experiments (section \ref{sec:exp}) show great potential, but also reveal some challenges.  These are discussed in the future work, followed by our conclusions.

\section{Time-of-Flight Camera}

\begin{figure}
  \vspace{-0.0cm}
  \centering
   {\epsfig{file = 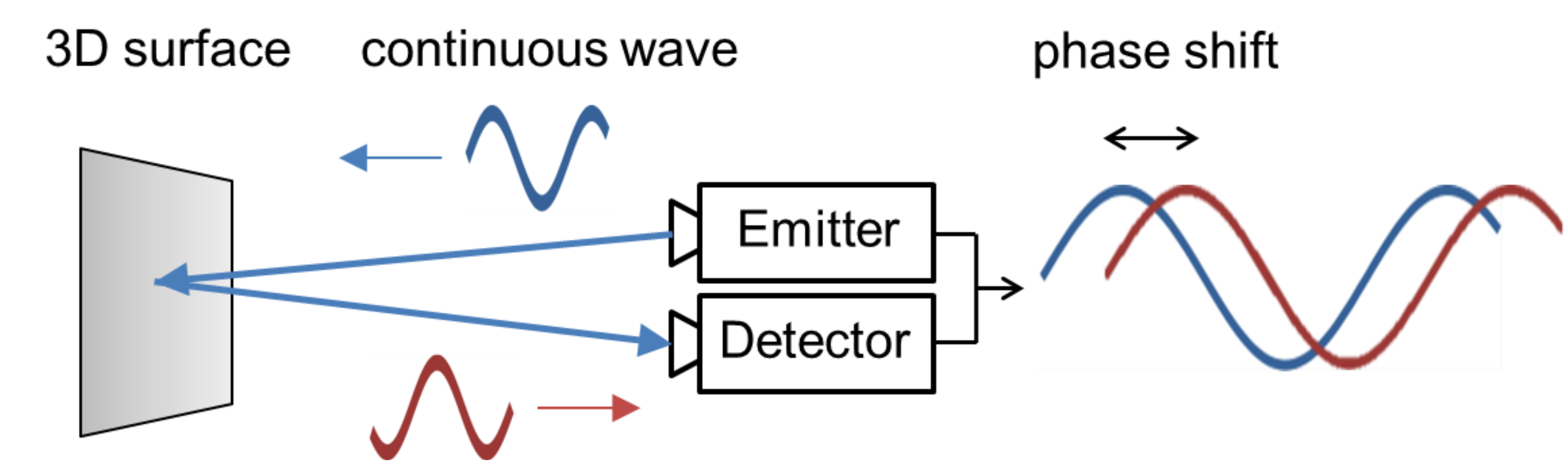, width = 7.5cm}}
  \caption{Time-of-Flight camera principle.}
  \label{fig:tof}
  \vspace{0cm}
\end{figure}

\begin{figure}
  \vspace{-0.0cm}
  \centering
   {\epsfig{file = 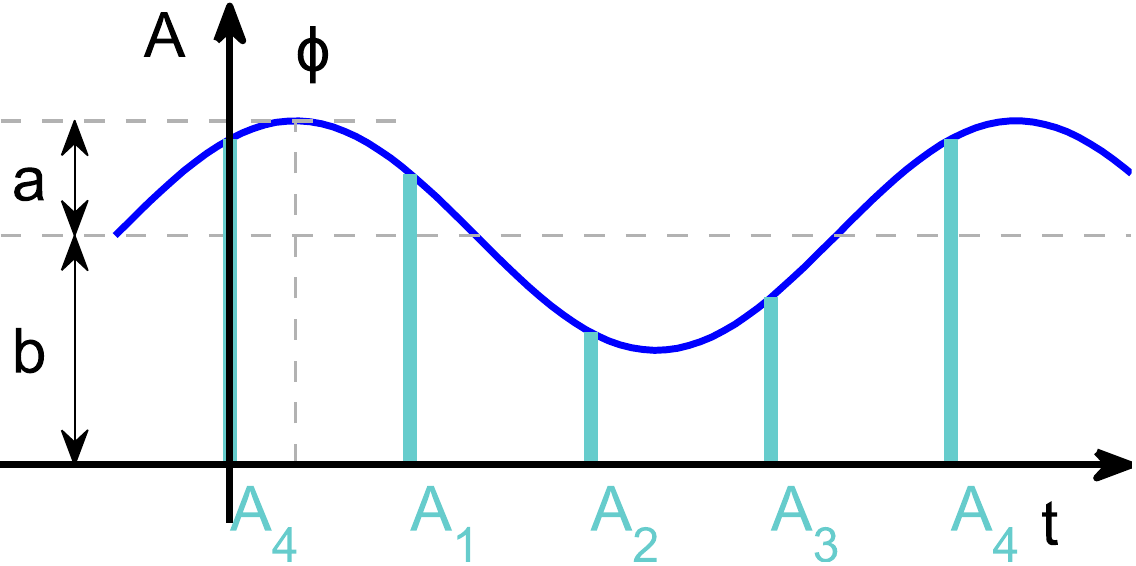, width = 5cm}}
  \caption{The reflected signal received by the TOF camera is sampled four times.  This allows to determine the signals phase shift $\varphi$.}
  \label{fig:tofsignal}
  \vspace{-0.1cm}
\end{figure}

\noindent A Time-of-Flight range camera is equipped with a near infrared light source (with a wavelength of about 850nm) that is modulated with a frequency of about 21 MHz (figure~\ref{fig:tof}).  The reflected light is collected onto a sensor capable of measuring the signal's phase ($\varphi$), amplitude (a) and offset (b) (figure~\ref{fig:tofsignal}, equations~\ref{eq:phase}-\ref{eq:amplitude}).  These are not measured directly, but can be determined using the four intensity measurements ($A_{1}$-$A_{4}$).  

\begin{equation}\label{eq:phase}
    \varphi = \arctan(\frac{A_{1}-A_{3}}{A_{2}-A_{4}}) + k \cdot 2 \pi
\end{equation}
\begin{equation}\label{eq:distance}
    D =\frac{c}{4\pi \cdot f_{mod}}  \cdot \varphi
\end{equation}
with c the speed of light in air and $f_{mod}$ the modulation frequency.
\begin{equation}\label{eq:offset}
    a = \frac{ \sqrt{ (A_{1}-A_{3})^{2} + (A_{2}-A_{4})^{2}  } }{2}
\end{equation}
\begin{equation}\label{eq:amplitude}
    b = \frac{A_{1}+A_{2}+A_{3}+A_{4}}{4}
\end{equation}

\noindent From the phase difference between the emitted and received signal, the total distance the light traveled is determined.  By dividing the total distance by two we obtain the object-camera distance (D).

\begin{figure}
  \vspace{-0.2cm}
  \centering
   {\epsfig{file = 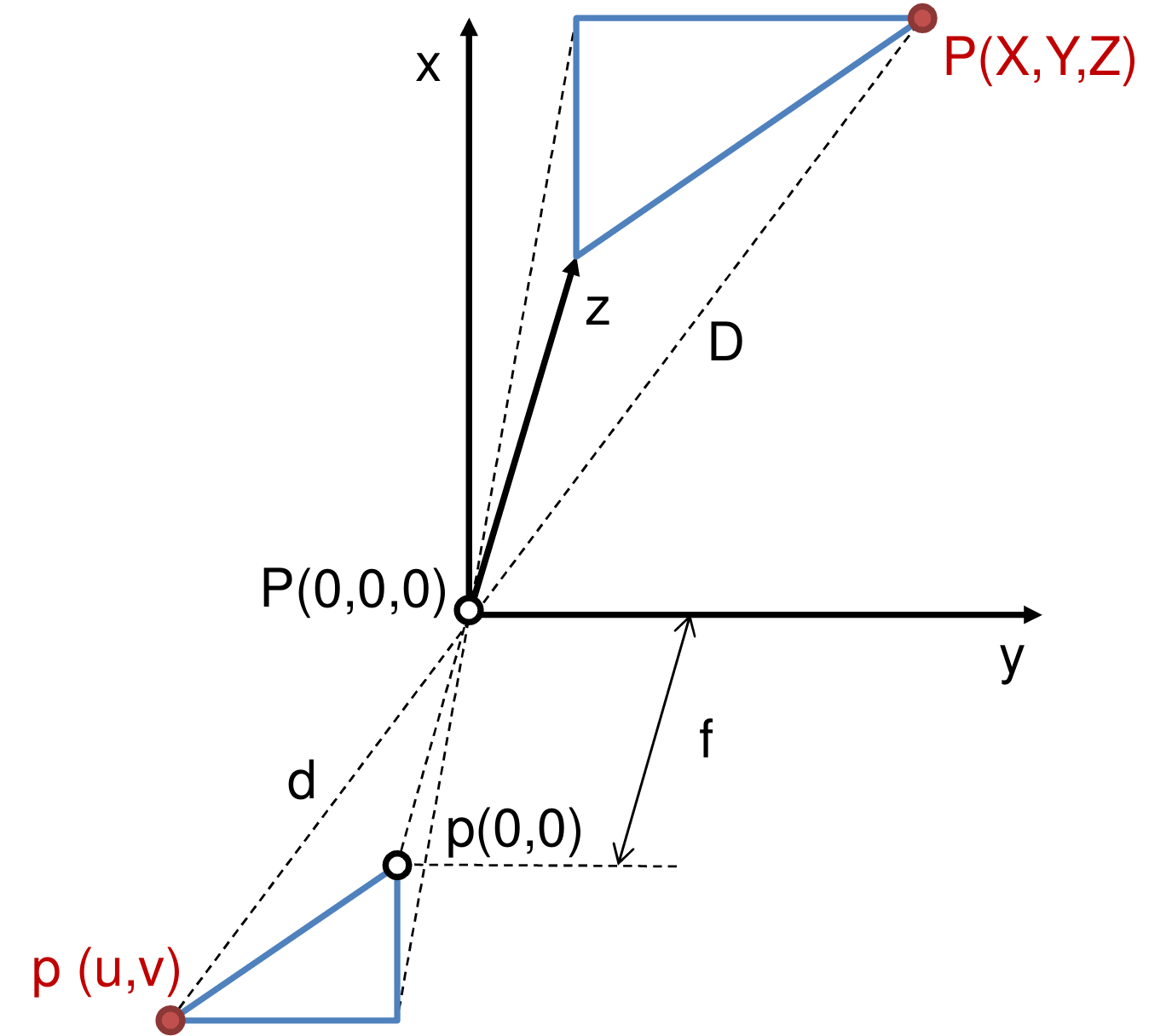, width = 5.5cm}}
  \caption{TOF camera pinhole model illustrating the projection of a point P(X,Y,Z) onto the TOF sensor: p(u,v).  The two highlighted triangles are of similar shape, hence the ratio of any two of their equivalent sides is equal.  This means that the known length $d$ and the measured distance $D$ suffice to determine the 3D coordinates of any point $p(u,v)$.}
  \label{fig:pinhole}
  \vspace{-0.1cm}
\end{figure}

If the TOF camera's focal length (f) is known we can calculate the 3D coordinates for every point.  These equations are easily deduced from figure \ref{fig:pinhole}.  

\begin{equation}\label{eq1}
    d = \sqrt{f^{2} + u^{2} + v^{2}}
\end{equation}
\begin{equation}\label{eq:XYZ}
	\begin{bmatrix} X \\ Y  \\ Z  \end{bmatrix}_{i} = \frac{D_{i} }{d} \begin{bmatrix} u \\ v \\ f \end{bmatrix}
\end{equation}

\noindent If radial distortion is present, this can be compensated by converting the distorted values (index d) to the undistorted values (index u).
\begin{equation}\label{eq:distortion}
	r_{u,i} = r_{d,i}+ k_{1} r_{d,i}^{3} + k_{2} r_{d,i}^{5}
\end{equation}
with
\begin{equation}\label{eq:distortion}
	r_{d,i} = \sqrt{u_{i}^{2}+v_{i}^{2}}
\end{equation}
We change the (u,v) coordinates accordingly:
\begin{equation}\label{eq:uvundist}
	\begin{bmatrix} u\\  v\end{bmatrix}_{u,i} = \frac{r_{u,i} }{r_{d,i}} \begin{bmatrix} u \\ v\end{bmatrix}_{d,i}
\end{equation}

\section{Thermal Infrared Camera}


\noindent All objects emit a certain amount of black body radiation as a function of their temperature. The higher an object's temperature, the more infrared radiation is emitted. A thermal infrared camera consist of an array of elements that measure this radiation.  They are typically sensitive in the far infrared range, at wavelengths of about 5-15$\mu m$.  Silicon or germanium lenses must be used, as glass does not transmit these wavelengths. 

Several types of IR sensor exist.  In general, a distinction can be made between cooled and uncooled infrared detectors.  We only consider the uncooled variety as they are cheaper and more compact.  In our experiments we use both the thermopile array and microbolometer type of detector.

In a thermopile array the heat radiated from an object is absorbed by a small membrane.  The temperature difference between the membrane and a thermal mass causes a difference in electric potential. This is known as the Seebeck effect.  This voltage can be converted to an absolute temperature measurement.

A microbolometer is a very similar device but instead of relying on the Seebeck effect, it uses the temperature coefficient of resistance. 

Both sensors are able to simultaneously measure a number of absolute temperatures, often visualized as a false color image.  While the microbolometer technology is more mature, the thermopile array is cheaper and smaller.  Although very limited at the moment (4x16px), an increase in resolution is expected which will soon make them an interesting alternative.

\section{\uppercase{Data fusion}}
\label{sec:fusion}

\begin{figure}
  \centering
   {\epsfig{file = 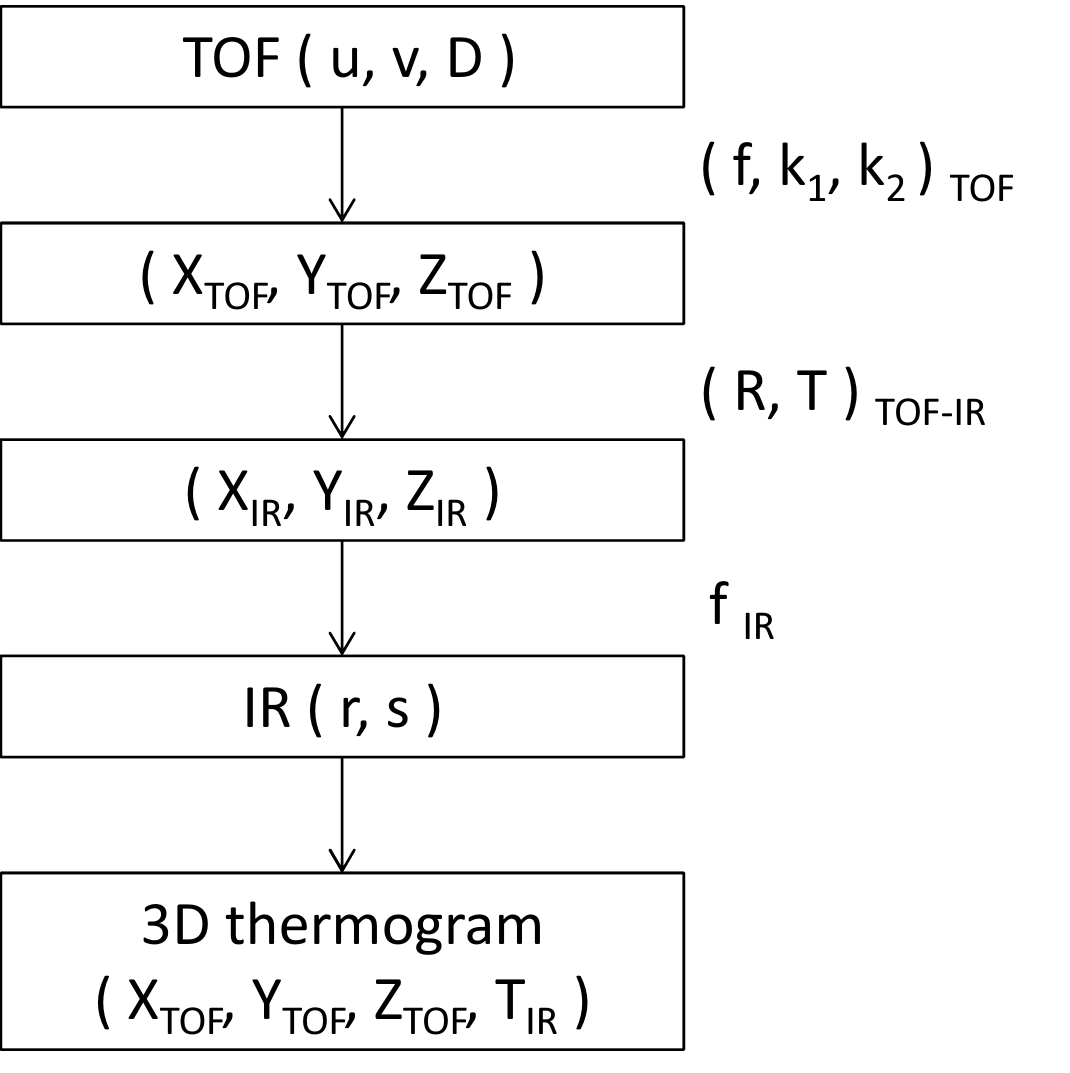, height= 5cm, trim=0 0 100 0}}
  \caption{In the data fusion process the range measurements are projected onto the IR sensor to obtain a 3D thermogram.}
  \label{fig:flow}
 \end{figure}

\noindent  To obtain the 3D thermogram we wish to use to detect people, the TOF and IR data must be fused.  A method to fuse TOF range data with images from a regular camera is proposed in \cite{Hanning2011}.  Their method relies on the cameras being perfectly parallel, a condition that is not met by our sensor.  The accuracy of the calibration also relies on the accuracy of the range measurements, which is typically quite low.

We propose a new data fusion algorithm enabling us to assign a temperature measurement to every range measurement.   Figure \ref{fig:flow} gives a graphical overview.  

\begin{figure}
  \centering
   {\epsfig{file = 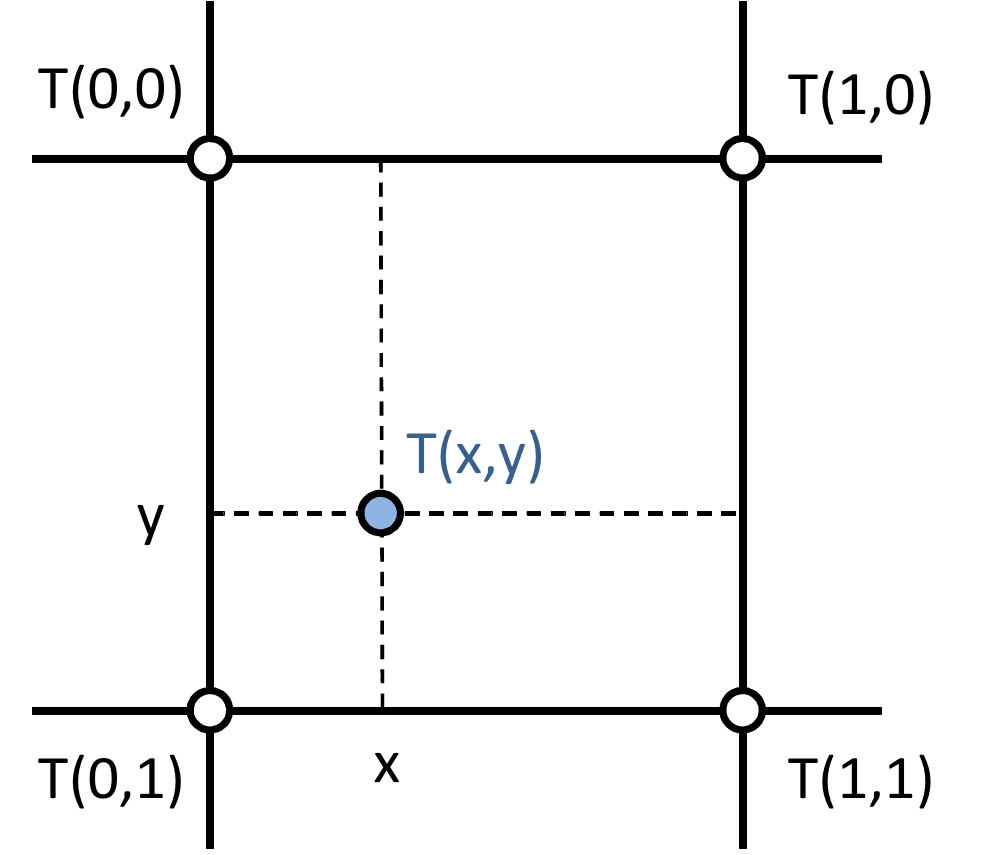, width = 4cm}}
  \caption{The point (x,y) is the projection of a point (X,Y,Z) onto the IR sensor.  The four other points represent the center of the four nearest pixels. We use bilinear interpolation to determine the temperature T(x,y).}
  \label{fig:interpolation}
 \end{figure}

 If the focal length of the TOF camera is known, every pixels' 3D coordinates can be calculated using their distance measurement (equation \ref{eq:XYZ}).  To obtain their 3D position in the IR camera reference frame we simply apply a translation and rotation to the 3D point cloud (equation \ref{eq:XYZTR}).  Projecting these points onto the calibrated IR sensor (equation \ref{eq:IRrs}) yields their position on the IR sensor from which we can obtain each 3D point's temperature measurement by bilinear interpolating between its four nearest neighbors (Figure \ref{fig:interpolation}, equation \ref{eq:bilinear}).

\begin{equation}\label{eq:XYZTR}
	\begin{bmatrix} X \\ Y  \\ Z  \end{bmatrix}_{IR} = \begin{bmatrix} R&|&T \end{bmatrix} \begin{bmatrix} X \\ Y  \\ Z  \\1 \end{bmatrix}_{TOF}
\end{equation}

\begin{equation}\label{eq:IRrs}
	Z \begin{bmatrix} r \\ s \\ 1  \end{bmatrix} = \begin{bmatrix} f_{IR} & 0 & 0 \\  0 & f_{IR} &0  \\ 0 & 0 & 1\end{bmatrix} \begin{bmatrix} X \\ Y  \\ Z \end{bmatrix}_{IR} 
\end{equation}

\small
\begin{equation}\label{eq:bilinear}
   T(x,y) =  \begin{bmatrix} 1-x & x \end{bmatrix}  \begin{bmatrix} T(0,0) & T(0,1) \\  T(1,0) & T(1,1) \end{bmatrix}  \begin{bmatrix} 1-y \\ y \end{bmatrix}
\end{equation}
\normalsize

\noindent Other interpolation methods may be investigated later on.

\section{\uppercase{Calibration}}
\label{sec:cal}

\noindent Various methods to calibrate the intrinsic parameters of both TOF and IR cameras exist.  A very popular method uses a simple planar target with a checkerboard pattern \cite{Zhang1999}.  For regular cameras, these calibration images also allow to determine their relative pose.  The problem in our case is that the two sensors do not share a common spectral sensitivity range and that both measure fundamentally different object properties.

A first part of the solution is to use the TOF cameras intensity measurements instead of the phase.  This eliminates the need for a complex three dimensional calibration target and problems with measurement errors in the range data.

\begin{figure}
  \vspace{-0.2cm}
  \centering
   {\epsfig{file = 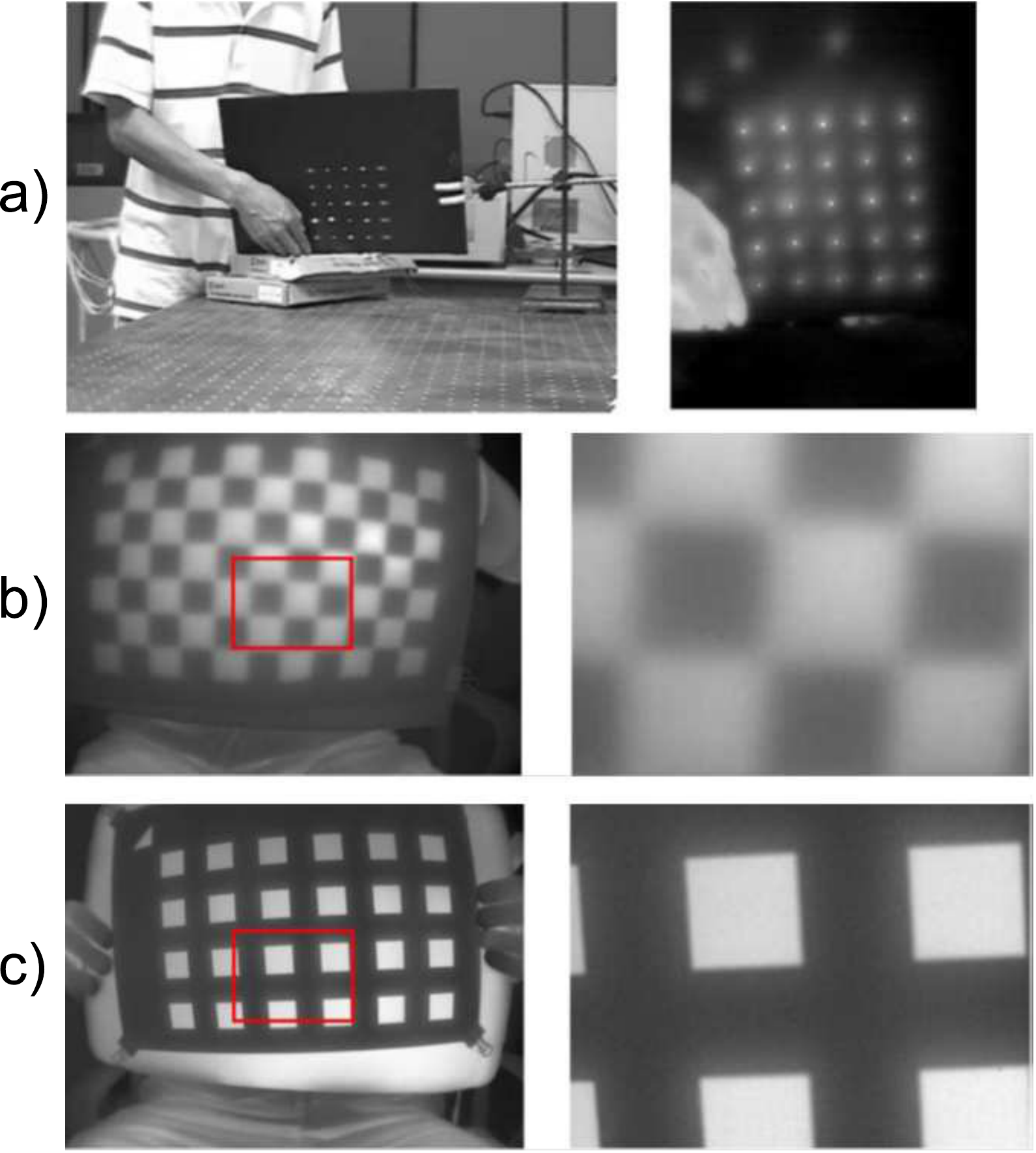, width = 7cm}}
  \caption{IR camera calibration targets suggested in literature.}
  \label{fig:chessboard}
  \vspace{-0.1cm}
\end{figure}

The second part of the solution is to use a calibration target that shows contrast in a very wide spectral range.  In the literature, a number of different solutions have been proposed. In \cite{Yang2011b} small light bulbs that emit light within the sensitivity range of the regular camera are used, the heat they generate can be measured by the IR camera (figure~\ref{fig:chessboard}a) . A more traditional 'checkerboard' calibration target (figure~\ref{fig:chessboard}b) is used in \cite{Vidas2012}.  By illuminating the pattern with a powerful lamp, the black regions will warm more quickly than the white regions because of the difference in absorption.  However, the temperature will soon even out and the contrast required to locate the corner points in the IR image accurately will quickly fade.  As a solution, they propose to use a mask with with regularly spaced holes (figure~\ref{fig:chessboard}c).  The mask is made of a material with contrasting colors to the background.  As a background, either a hotter or colder object is used, providing the required contrast in the IR image.  As both materials can be thermally insulated (e.g. by air), this approach doesn't suffer from the fading as much.  The mask however must be designed very carefully.  The inside edges must be made 'infinitely' thin so that the 'back edge' does not interfere with the measurement when the mask is viewed at an angle.  Also, the mask must be sufficiently rigid to assure it remains perfectly planar.

A method to calibrate a structured light and IR camera system is proposed in \cite{Yang2011}.  Their method requires higher resolutions sensors and a large amount of calibration measurements, as they estimate a large number of parameters in a single optimization procedure.

In a first step the calibration of the intrinsic camera parameters of the TOF and IR cameras is performed.  This is followed by the calibration of the relative pose of the two cameras.  As mentioned before, the 3D translation between the two cameras is assumed to be known.  This is a reasonable assumption as we can measure their position sufficiently accurately on the printed circuit board they are mounted on.  Also, a small error in the translation parameters will result in a small error in the pixel mapping.

The only parameters that remain to be estimated are the relative orientation of the two cameras.  To do this we measure objects at a known distance (this avoids relying on the distance measurement precision of the TOF camera) using the TOF camera.  As we already performed the intrinsic calibration, the object's image and its distance allow us to calculate its 3D position in the world ordinate system.  An initial estimate of the rotation and the known translation are used to calculate the points' position in the IR camera ordinate system.  The projected point will not coincide perfectly with the actual measurement point in the IR image.  This projection error is minimized in order to find the optimal rotation matrix R.  As the error function we use the euclidean distance between the virtual, projected point, and the real measurement.

\begin{equation}\label{eq1}
   E_{i} = |P_{m, i}(u_{i}, v_{i}, D_{i}) - P_{p, i}(u_{i}, v_{i}, D_{i}, R)|
\end{equation}

\begin{equation}\label{eq1}
   R = arg min(\sum_{i} E_{i})
\end{equation}

\noindent To find the object's position in the TOF and IR images, we may simply pick the pixel with the highest intensity, or highest temperature respectively.  Due to the low resolution this would result in a significant error.  In order to achieve sub-pixel precision, we can fit a function to the local neighborhood of this pixel, and determine the position corresponding to the maximum of this function.

\section{\uppercase{Preliminary Experiments}}
\label{sec:exp}

\begin{figure}
  \vspace{-0.2cm}
  \centering
   {\epsfig{file = 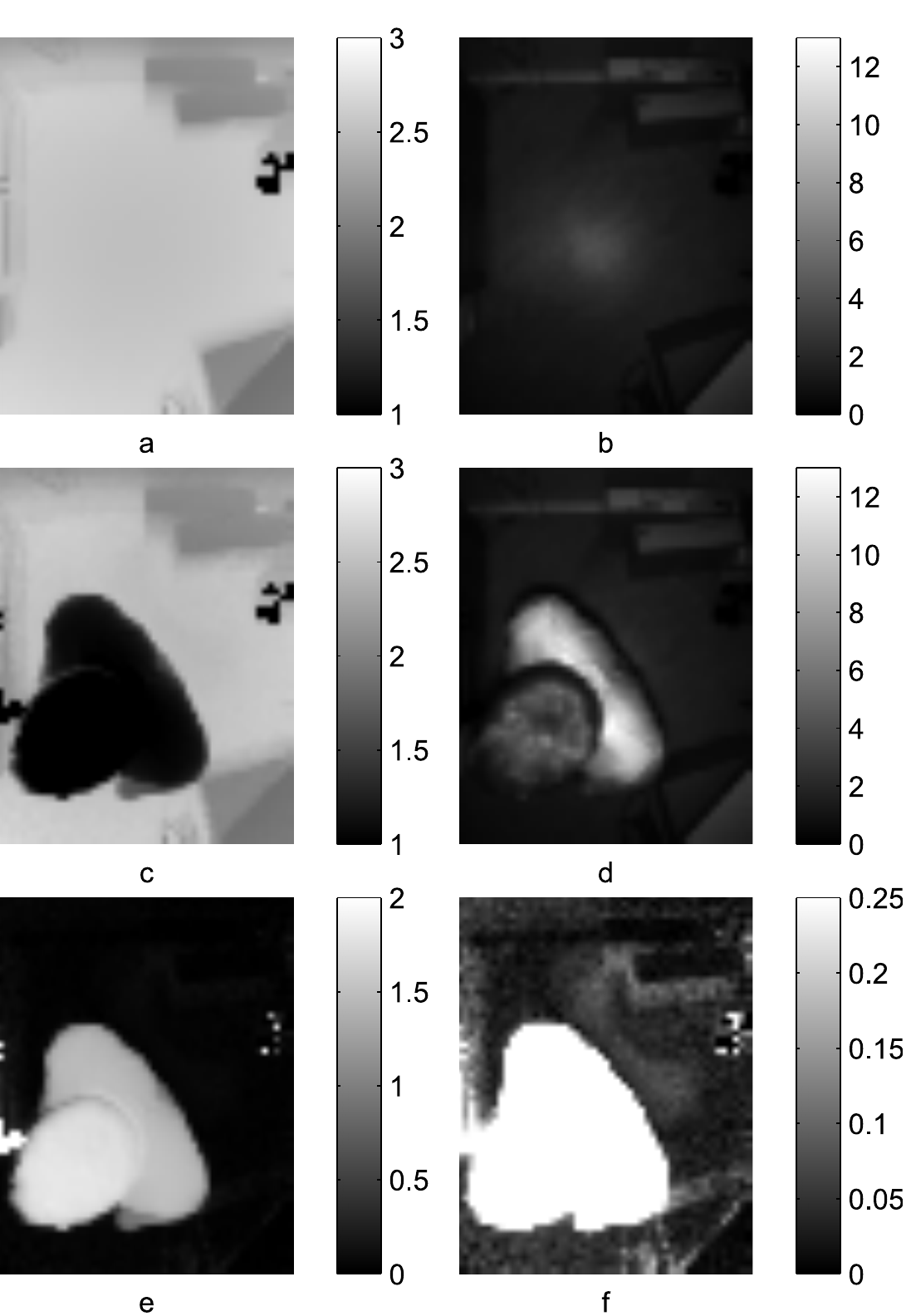, width = 7.5cm}}
  \caption{ a) Average distance (background). b) Average intensity. c) Single distance image. d) Single intensity image. e) Background subtracted, absolute difference of a and c. f) Background subtracted, with rescaled range.  The data contain some outliers due to over or underexposure.}
  \label{fig:tofdata}
  \vspace{-0.1cm}
\end{figure}

\begin{figure}
  \vspace{-0cm}
  \centering
   {\epsfig{file =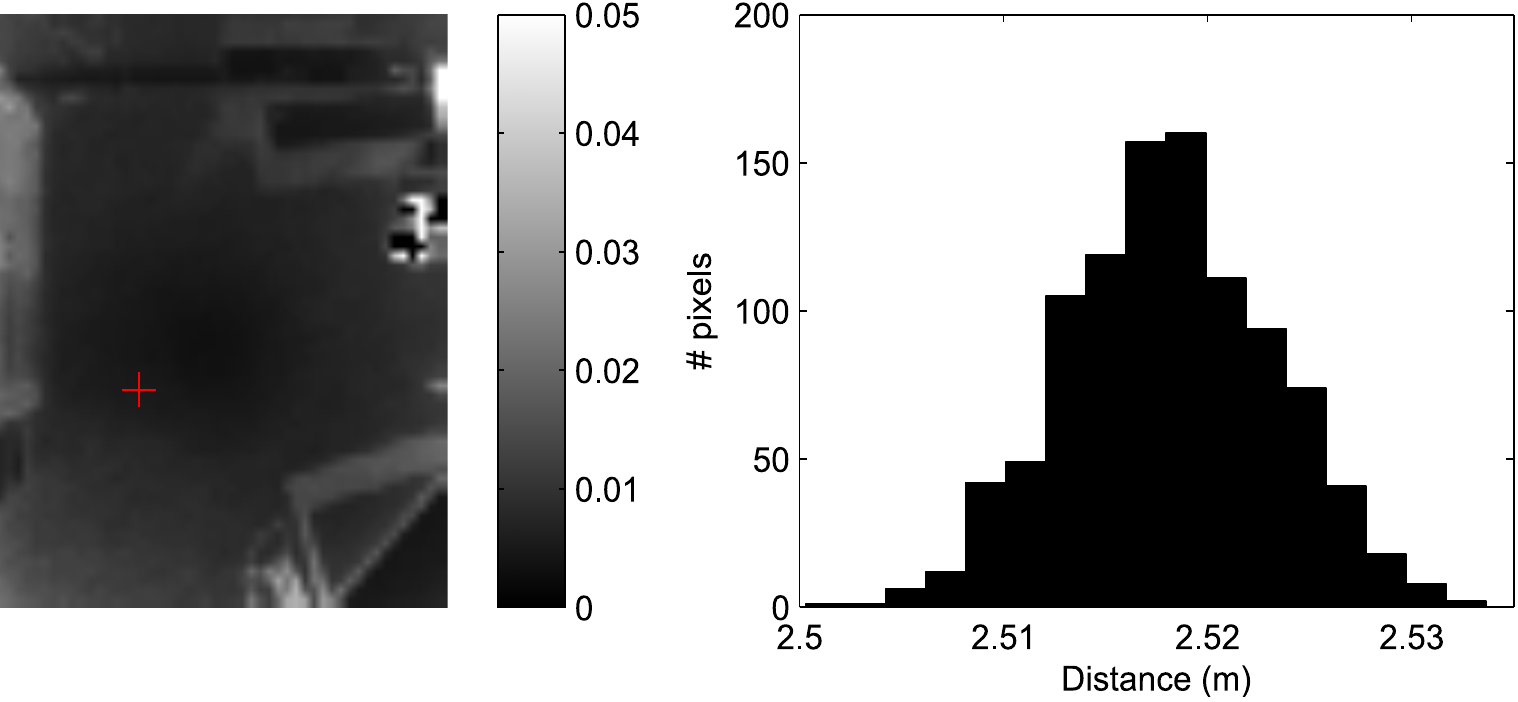, width = 7.5cm}}
  \caption{Left: Standard deviation (in meters) of range measurements. Right: Histogram showing the distribution of the measured distances of the highlighted pixel.}
  \label{fig:std}
  \vspace{-0.1cm}
\end{figure}

\noindent The experiments are performed with an IFM O3D201 TOF camera with a resolution of 64x50 pixels and at a modulation frequency of 21MHz.  The thermal camera used is the Flir E40bx microbolometer, which has a resolution of 160x120 pixels.

In a first experiment the TOF camera was fixed to the ceiling of a room and recorded a sequence of images.  A set of 1000 consecutive frames without moving objects was averaged and used as 'background' (figure~\ref{fig:tofdata}a).  The background was subtracted from the another frame  (figure~\ref{fig:tofdata}c) in an attempt to segment moving objects (figure~\ref{fig:tofdata}e,f).

As can be seen in figure \ref{fig:tofdata}f, our measurements are subject to noise.  The noise on the phase measurement, results in a distance error.  The standard deviation per pixel in the 1000 background frames is shown in figure \ref{fig:std}, next to the histogram of the range measurement of the highlighted pixel.  The average error is typically about one percent, which is acceptable for many applications.  The noise level depends on a variety of properties such as illumination power, object reflectivity, surface orientation, distance, etc.  In general the error can be modeled (e.g. using gaussian mixture models) quite well.  This allows the probability a pixel belongs to the background or foreground to be calculated.

\begin{figure}
  \vspace{-0.2cm}
  \centering
   {\epsfig{file = 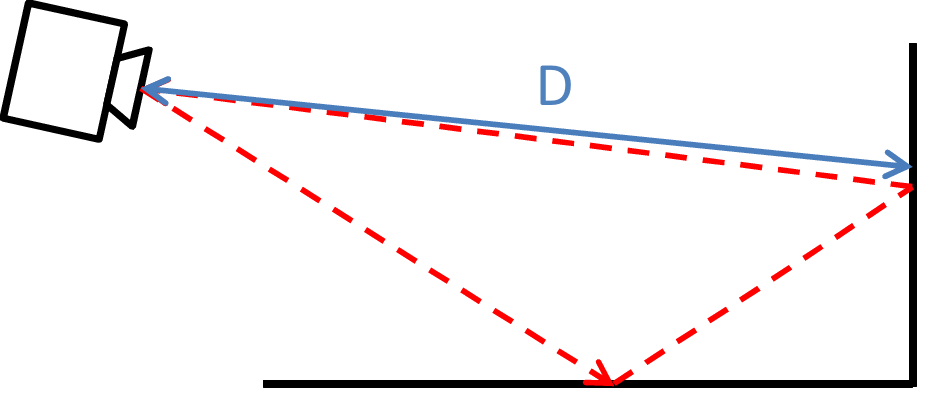, width = 4cm}}
  \caption{The direct reflection off the wall (blue, solid line) provides the correct range measurement D.  This measurement, however, will be corrupted by other signals (e.g. the red, dotted line), which will have a larger phase shift as they have traveled a longer distance.  This is known as the multi-path interference error.}
  \label{fig:multipath}
  \vspace{-0.1cm}
\end{figure}

\begin{figure}
  \vspace{-0.2cm}
  \centering
   {\epsfig{file = 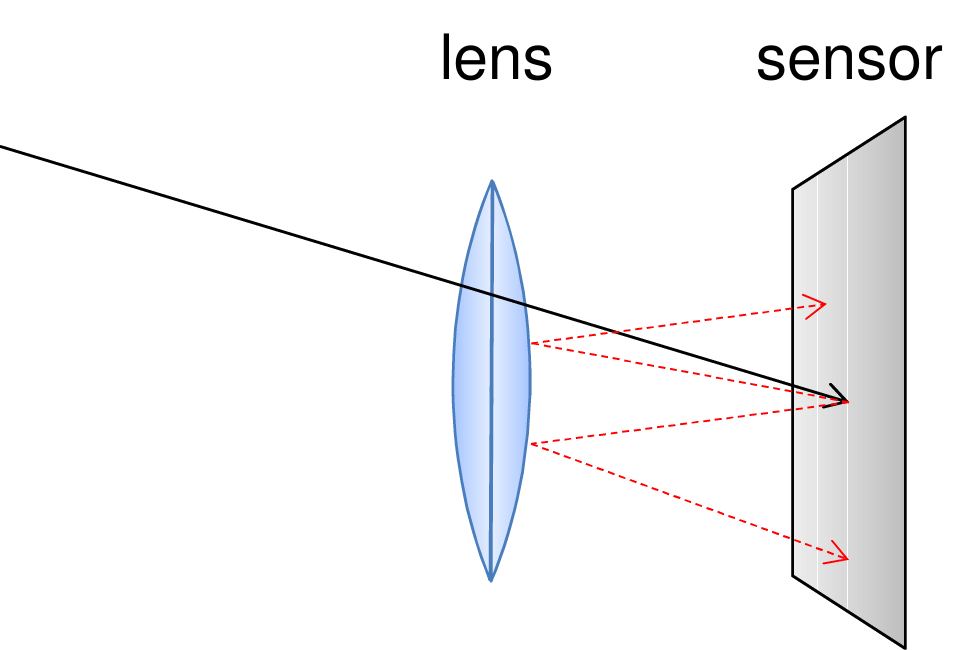, width = 5cm}}
  \caption{The scattering effect of a ray of light in a TOF camera.  The incoming ray is not completely absorbed by the pixel but partially scattered.  Part of the scattered energy reflects on the lens surface and back onto the sensor, disturbing the measurements.}
  \label{fig:scattering}
  \vspace{-0.1cm}
\end{figure}

However, if we carefully compare the image with subtracted background (figure~\ref{fig:tofdata}f) to the standard deviations in figure \ref{fig:std}, we see that some of the absolute differences on the background are substantially greater than the expected noise.  These errors are due to multi-path interference (figure~\ref{fig:multipath}) and scattering (figure~\ref{fig:scattering}).

\section{\uppercase{Future work}}
\label{sec:future}

\noindent  While the current hardware setup has a small field of view, a prototype with a larger field of view is being developed.  This will allow to monitor a reasonably large volume.  The calibration routine also allows to fuse data from multiple sensors to extend it even more.

We intend to integrate the calibration of the intrinsic and extrinsic parameters of the combined sensor into one automatic procedure.  To determine the point spread functions for both the IR and TOF cameras, a system allowing to systematically repositioning the calibration target using a robot will be set up.  The generated data will provide a better understanding of the multi-path interference and scattering errors observed in our experiments.  Error compensation methods such as \cite{Karel2012}  \cite{Mure2007} allow to improve the reliability in segmenting moving objects and will increase correspondence accuracy between the IR and range data.

A set of fused range and IR data will be generated and used to train people detection algorithms.  The three main hypotheses mentioned in the introduction will be thoroughly tested.  Applying 3D tracking algorithms will increase the robustness and enable the system to cope with occlusion.

\section{\uppercase{Conclusions}}
\label{sec:conclusion}

\noindent A combined sensor for the detection of people using fused geometric and infrared radiation data was introduced.  We explained the working principles of both sensors and illustrated and addressed some important accuracy issues that arose during experiments.  A method to calibrate a system with known relative position and unknown relative orientation was proposed.  Three key areas in people detection that could benefit greatly from the fused IR and range data were determined and will be investigated in future work.

\section*{\uppercase{Acknowledgments}}
\label{sec:ack}

\noindent We thank ICRealisations for their input and for providing a prototype system for experimental work, and Xenics for providing a thermal infrared camera for verification purposes.

\bibliographystyle{apalike}
{\small
\bibliography{WAB_ICINCO2013}}

\vfill
\end{document}